\documentclass[sigconf]{acmart}
\usepackage{algorithm}
\usepackage{algorithmic}
\usepackage[most]{tcolorbox}

\AtBeginDocument{%
  }

\setcopyright{acmlicensed}
\copyrightyear{2025}
\acmYear{2025}
\acmDOI{}
\acmConference[KDD '25]{31st ACM SIGKDD Conference on Knowledge Discovery and Data Mining}{August 03--07, 2025}{Toronto, Canada}
\acmISBN{}

\begin{document}

\title{Multi-Stage Verification-Centric Framework for \\ Mitigating Hallucination in Multi-Modal RAG}

\author{Baiyu Chen}
\authornote{Both authors contributed equally to this research.}
\orcid{0009-0000-8617-9635}
\affiliation{%
  \institution{The University of New South Wales}
  \city{Sydney}
  \state{NSW}
  \country{Australia}
}
\email{breeze.chen@student.unsw.edu.au}

\author{Wilson Wongso}
\authornotemark[1]
\affiliation{%
  \institution{The University of New South Wales}
  \city{Sydney}
  \state{NSW}
  \country{Australia}
}
\email{w.wongso@unsw.edu.au}

\author{Xiaoqian Hu}
\affiliation{%
  \institution{The University of New South Wales}
  \city{Sydney}
  \state{NSW}
  \country{Australia}
}
\email{xiaoqian.hu@student.unsw.edu.au}

\author{Yue Tan}
\affiliation{%
  \institution{The University of New South Wales}
  \city{Sydney}
  \state{NSW}
  \country{Australia}
}
\email{yue.tan@unsw.edu.au}

\author{Flora Salim}
\affiliation{%
  \institution{The University of New South Wales}
  \city{Sydney}
  \state{NSW}
  \country{Australia}
}
\email{flora.salim@unsw.edu.au}

\renewcommand{\shortauthors}{Chen et al.}


\begin{abstract}
This paper presents the technical solution developed by team CRUISE for the KDD Cup 2025 Meta Comprehensive RAG Benchmark for Multi-modal, Multi-turn (CRAG-MM) challenge. The challenge aims to address a critical limitation of modern Vision Language Models (VLMs): their propensity to hallucinate, especially when faced with egocentric imagery, long-tail entities, and complex, multi-hop questions. This issue is particularly problematic in real-world applications where users pose fact-seeking queries that demand high factual accuracy across diverse modalities. To tackle this, we propose a robust, multi-stage framework that prioritizes factual accuracy and truthfulness over completeness. Our solution integrates a lightweight query router for efficiency, a query-aware retrieval and summarization pipeline, a dual-pathways generation and a post-hoc verification. This conservative strategy is designed to minimize hallucinations, which incur a severe penalty in the competition's scoring metric. Our approach achieved 3rd place in Task 1, demonstrating the effectiveness of prioritizing answer reliability in complex multi-modal RAG systems. Our implementation is available at \url{https://github.com/Breezelled/KDD-Cup-2025-Meta-CRAG-MM}.
\end{abstract}




\maketitle

\section{Introduction}

Recent progress in Vision-Language Models (VLMs) has enabled richer multimodal reasoning, allowing systems to interpret and generate responses from visual and textual inputs. These models have demonstrated impressive capabilities in tasks such as visual question answering, image captioning. However, despite these advances, VLMs still struggle with factual accuracy, frequently producing hallucinations, confident but incorrect or fabricated outputs that are not grounded in the input image or external knowledge~\cite{huang2025survey}. This challenge is particularly acute for real-world applications like smart glasses, which process a continuous stream of dynamic, egocentric, and often low-quality visual data. From this first-person perspective, user queries are not only complex but also frequently vague or ambiguous. They often require real-time information retrieval (e.g., "what is the cost of this product on Amazon?"), involve pronouns that need visual grounding (e.g., "does this contain sodium?"), and depend on external knowledge for verification. These factors compound the risk of hallucination, demanding systems that are not only capable but fundamentally reliable.

The Retrieval-Augmented Generation (RAG) paradigm offers a promising solution by grounding model responses in external knowledge~\cite{gao2023retrieval}. However, extending RAG to the multi-modal setting (MM-RAG) introduces significant challenges, including efficient retrieval, deciding whether and when to query external sources, multi-source synthesis, and maintaining conversational context. These challenges arise from the increased complexity of integrating heterogeneous modalities such as images, user queries, retrieved web passages, and structured knowledge graph entries. Unlike traditional text-based RAG, MM-RAG must reconcile visual inputs with both unstructured and structured textual sources, while maintaining referential grounding and contextual coherence across modalities. To evaluate these challenges in real-world conditions, the KDD Cup 2025 Meta CRAG-MM (Comprehensive RAG Benchmark for Multi-modal Multi-turn Question Answering)~\cite{wang2025crag} introduces a rigorous benchmark designed to foster innovation in MM-RAG systems.
To address the compounded challenges of factual grounding and response reliability in MM-RAG, we propose a verification-centric framework designed to prioritize accuracy over completeness. Our approach incorporates a lightweight query router to identify questions requiring real-time or external information, a multi-source retrieval module combining web and knowledge graph search with reranking and dynamic thresholding, a dual-path generation strategy to balance prior knowledge and grounded context, and a structured verification mechanism that enforces factual consistency through self-consistency and chain-of-verification protocols. In summary, our key contributions are as follows:
\begin{itemize}
    \item We propose a multi-stage, verification-centric framework for MM-RAG that explicitly prioritizes factual accuracy.
    \item We detail the implementation of key components, including a lightweight query router for efficiency, a query-aware retrieval pipeline with reranking and dynamic thresholding for relevance, and a novel dual-path generation with a self-consistency check and chain-of-verification for robust answer validation.
    \item We demonstrate the effectiveness of our system on the KDD Cup 2025 Meta CRAG-MM challenge, and release our implementation to support reproducibility and future research on building reliable MM-RAG systems that reduce hallucination in real-world, egocentric scenarios.
\end{itemize}
\section{Benchmark and Task Definition}
\label{sec:task-def}

The CRAG-MM challenge is designed as a comprehensive testbed for MM-RAG systems, particularly for egocentric and wearable device scenarios. This section details the benchmark's components, tasks, and the formal problem definition.

\subsection{Benchmark Overview}

The CRAG-MM benchmark is designed to evaluate the capabilities of MM-RAG systems in egocentric and real-world scenarios. It comprises three key components:

\textbf{(1) Image Set}: Consists of 5,000 images, including 3,000 egocentric images captured using Ray-Ban Meta smart glasses and 2,000 third-person images collected from public sources. This mix simulates realistic inputs encountered by wearable AI systems across 13 domains such as \textit{Books}, \textit{Food}, \textit{Math \& Science}, \textit{Shopping}, \textit{Animals}, and \textit{Vehicles}.

\textbf{(2) QA Set}: Includes over 5,000 question-answer pairs spanning the same 13 domains. Questions are categorized into four types:
    \begin{itemize}
        \item \textbf{Simple Questions}: Require either direct visual recognition or a single piece of external knowledge.
        \item \textbf{Multi-hop Questions}: Involve reasoning over multiple pieces of information to derive an answer.
        \item \textbf{Comparison and Aggregation Questions}: Require synthesis and comparison of several data points.
        \item \textbf{Reasoning Questions}: Target implicit knowledge or logical inference beyond direct retrieval.
    \end{itemize}
    Questions appear in both single-turn and multi-turn formats, enabling assessment of stateful contextual reasoning.

\textbf{(3) Retrieval Contents}:
    \begin{itemize}
        \item \textbf{Image Search API}: A mock API that takes an image as input and returns visually similar images with structured metadata (e.g., description, caption, summary) from a mock knowledge graph.
        \item \textbf{Text-Based Web Search API}: Accepts text queries and returns relevant web pages, including titles, URLs, snippets and last updated time.
    \end{itemize}
Both APIs are augmented with hard negatives to simulate noise, ambiguity, and real-world retrieval complexity.

Together, these components enable comprehensive assessment of the ability of MM-RAG systems to ground responses, synthesize multi-source information, and maintain factual consistency in dynamic, egocentric scenarios.

\subsection{Task Definition}
The competition is structured into three complex tasks: \\
\textbf{Task 1: Single-source Augmentation}: Tests the basic RAG capability using a provided image-based mock Knowledge Graph (KG) as the external source. \\
\textbf{Task 2: Multi-source Augmentation}: Increases complexity by adding a web search API, requiring systems to synthesize information from both the KG and noisy web pages. \\
\textbf{Task 3: Multi-turn QA}: Focuses on context understanding and maintaining coherent, stateful conversations over 2-6 turns, where later turns may or may not depend on the initial image.

Together, the three tasks progressively evaluate different dimensions of multimodal RAG systems under realistic, open-ended QA scenarios. Task 1 serves as a foundation, assessing the system’s ability to retrieve and ground answers using a single, structured source (KG) tied to the input image. Task 2 introduces additional challenges by requiring systems to retrieve from multiple heterogeneous sources, thereby testing their robustness to noisy or redundant information. Task 3 emphasizes conversational consistency and memory, evaluating whether a system can maintain contextual grounding and answer coherently across multiple turns. This tiered structure enables a holistic assessment of factuality, retrieval quality, and dialogue-level reasoning in multimodal QA.

\subsection{Evaluation}
The challenge's most defining feature is its rigorous evaluation protocol, which combines automated metrics with manual review and is subject to strict performance constraints.
\textbf{Scoring Metric.} Responses are scored on a four-level scale: \textit{Perfect} (1.0) for fully correct and factual answers; \textit{Acceptable} (0.5) for useful answers with minor, non-harmful errors; \textit{Missing} (0.0) for abstentions (e.g., ``I don't know''); and \textit{Incorrect} (-1.0) for factually wrong or hallucinatory answers. This scoring system imposes a strong incentive for systems to avoid speculation, as a single hallucinated answer cancels out the benefit of a correct one. Our framework is therefore fundamentally guided by the goal of maximizing this truthfulness-focused score.
\textbf{Evaluation Process and Constraints.} During the competition, automated evaluation is used for leaderboard scoring, with only the first 75 BPE tokens of a response being considered. However, final rankings are determined by manual review of the top-10 teams, where human annotators assess the quality and factuality of the full response. This process is governed by strict performance constraints: a 10-second timeout is enforced per conversational turn, and any submission exceeding this limit is considered a failure. Furthermore, systems must handle time-sensitive queries, as the ground truth reflects correctness at the time of data collection.

\begin{figure*}[h]
    \centering
    \includegraphics[width=1\textwidth]{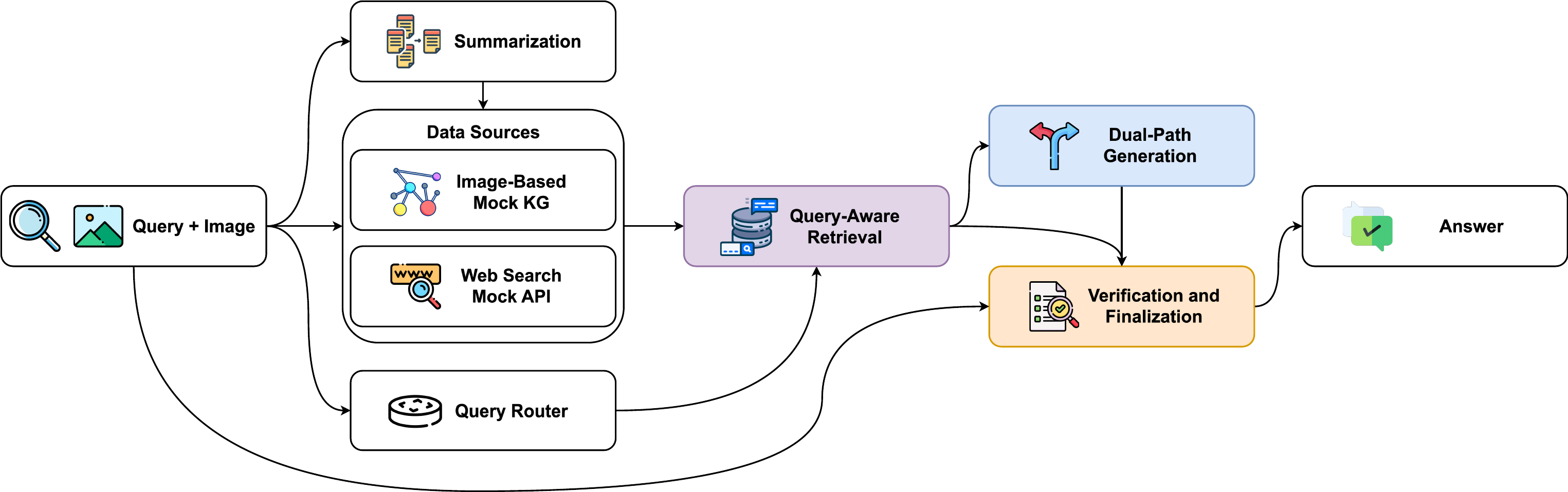}
    \caption{The overall pipeline of our solution.}
    \label{fig:pipeline}
\end{figure*}
\section{Methodology}

To address the challenges outlined in Section \ref{sec:task-def}, we designed a multi-stage, verification-centric pipeline that prioritizes factual accuracy and response reliability. Our framework, illustrated in Figure \ref{fig:pipeline}, processes each query through four key stages: (1) query routing, (2) query-aware retrieval and summarization, (3) response generation via dual pathways, and (4) post-hoc verification using a combination of confidence estimation and self-consistency. This solution is applied consistently across all three CRAG-MM tasks.

\subsection{Lightweight Query Routing}
In real-world multimodal QA systems, especially those deployed on latency-critical platforms such as smart glasses, efficient query routing is essential. Not all user questions warrant the same computational treatment: some can be answered directly from visual input, while others require external or time-sensitive knowledge. Applying a full RAG pipeline to every query introduces unnecessary latency and resource usage. Prior work has shown that when response latency exceeds 7–10 seconds, users report feeling significantly more tense, tired, frustrated, and sluggish, all of which contribute to a worse subjective user experience~\cite{10.1145/2600428.2609627}. These findings are particularly relevant to the CRAG-MM challenge, which enforces a strict 10-second response limit per turn. Moreover, CRAG-MM includes a diverse range of question types—from simple image-based recognition to temporal queries, making adaptive query handling even more critical.

To address this, we design a lightweight query routing module as the first stage of our pipeline, enabling early classification of queries to avoid unnecessary computation. We employ a compact instruction-tuned model, LLaMA-3.2-1B-Instruct\footnote{\url{https://huggingface.co/meta-llama/Llama-3.2-1B-Instruct}}~\cite{grattafiori2024llama}, as a dedicated router. This smaller model is computationally efficient and can semantically classify queries with minimal overhead.
A key design consideration is the well-documented tendency of large language models to generate hallucinated or outdated answers when queried about facts or recent events beyond their training cutoff. To mitigate these risks and optimize downstream decisions, our router explicitly predicts two binary attributes for each question \( q_i \in Q \):

\begin{itemize}
    \item \textbf{Needs External Info} \( e \in \{0,1\} \): whether answering \( q_i \) requires knowledge beyond the input image. This helps avoid hallucinations caused by over-reliance on model priors when the necessary information is not visible.
    \item \textbf{Is Real-Time} \( r \in \{0,1\} \): whether \( q_i \) is time-sensitive (e.g., involving "today", "latest", etc.). This is aimed at reducing the risk of outdated model knowledge when responding to temporal queries.
\end{itemize}

The classification is performed by prompting the router with a structured template, and the resulting decision \( D = \{e, r\} \) informs subsequent processing steps. If \( e = 0 \), the retrieval stage is skipped, the process then moves directly to dual-path generation (Section \ref{sec:dual-gen}), where the context is treated as empty, conserving computational resources and improving latency. Full prompt details are provided in Appendix~\ref{appendix:routing}.

\begin{figure}[h]
    \centering
    \includegraphics[width=0.45\textwidth]{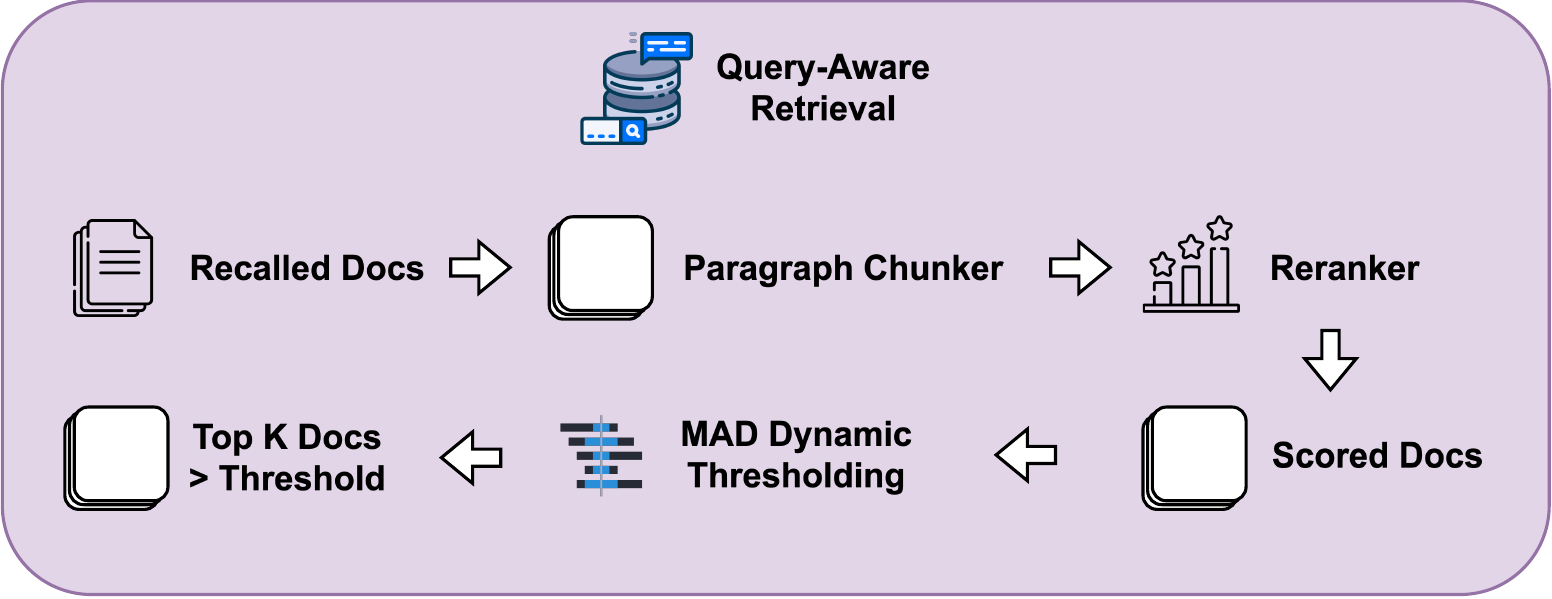}
    \caption{The detail of Query-Aware Retrieval module.}
    \label{fig:retrieval}
\end{figure}

\subsection{Query-Aware Retrieval}
Traditional RAG methods often rely on static image captions, raw user query for retrieval. These approaches are insufficient under the CRAG-MM setting, where queries are often ambiguous and the knowledge sources are heterogeneous and noisy. As a result, traditional RAG tends to retrieve irrelevant or weakly grounded contexts. To improve grounding quality and reduce noisy retrievals, we design a query-aware retrieval pipeline that explicitly incorporates the query intent into the visual summarization and passage scoring process. An overview of this module is shown in Figure~\ref{fig:retrieval}.

For each query \(q_i\), we first generate a query-aware image summary \(s_i\) using LLaMA-3.2-11B-Vision-Instruct\footnote{\url{https://huggingface.co/meta-llama/Llama-3.2-11B-Vision-Instruct}}~\cite{grattafiori2024llama}, an instruction-tuned vision-language model capable of producing grounded and context-sensitive descriptions:
\begin{equation}
s_i = f_(q_i, I_i),
\end{equation}
where \(I_i\) is the input image and \(f\) denotes the vision language model. The prompt of summarization is provided in Appendix \ref{appendix:summarizing} We then form the retrieval query by concatenating the textual question and the summary:
\begin{equation}
q_i' = q_i \;\Vert\; s_i.
\end{equation}

\textbf{Initial Recall.} Depending on the routing decision \(e\), we perform either text-based or image-based retrieval with a unified search pipeline \(R\) provided by organizer:
\begin{equation}
D_i =
\begin{cases}
R(q_i', K_{\text{recall}}), & \text{if web search} \\
R(I_i, K_{\text{recall}}), & \text{if image search}
\end{cases}
\quad
D_i = \{d_{ij}\}_{j=1}^{K_{\text{recall}}}
\end{equation}
where \(K_{recall}\) is top K recall and \(K_{recall} = 10\) in our implementation. Each retrieved item \( d_{ij} \) contains various structured attributes. The search pipeline was carried out using two modality-specific embedding models: bge-large-en-v1.5\footnote{\url{https://huggingface.co/BAAI/bge-large-en-v1.5}} for text-based queries and clip-vit-large-patch14-336\footnote{\url{https://huggingface.co/openai/clip-vit-large-patch14-336}}~\cite{radford2021learning} for image-based queries. We specifically extract the \verb|description|, \verb|caption|, and \verb|summary| fields from the mock knowledge graph (KG) and web snippets. These contents are then normalized into individual paragraphs by splitting on newlines. The resulting paragraph-level chunks are scored independently during reranking. Chunking is essential, as the retrieved fields may span entire documents, often exceeding the reranker’s maximum input sequence length. In practice, we found paragraph-level granularity to be well-suited for most modern reranker models.


\textbf{Reranking with Dynamic Thresholding.} To improve retrieval precision, we rerank the normalized paragraph chunks using bge-reranker-v2-m3\footnote{\url{https://huggingface.co/BAAI/bge-reranker-v2-m3}}~\cite{chen-etal-2024-m3}, a state-of-the-art multilingual cross-encoder. Each chunked paragraph from \( d_{ij} \) is scored against the expanded query \( q_i' \) as follows:
\begin{equation}
score = f_{\text{rank}}(q_i', chunk),
\end{equation}
where \( f_{\text{rank}} \) denotes the reranker and \( score \in [0,1] \) is the predicted relevance score. Top-ranked chunks are retained for generation based on a dynamic threshold.

A significant challenge in RAG is handling the noisy, long-tailed nature of retrieved documents. To address this, we move beyond conventional fixed-threshold filtering and utilize a principled, statistics-based dynamic cutoff. We leverage the Median Absolute Deviation (MAD) to suppress noisy evidence while retaining salient chunks. MAD is a robust statistic widely used to measure the variability of a distribution, particularly effective under skewed or heavy-tailed distributions where traditional metrics like standard deviation may be unreliable. Given that reranker scores are often unevenly distributed with a small number of highly relevant chunks and many low-scoring distractors, MAD offers a principled and stable way to estimate a threshold without being overly sensitive to outliers. Let \(\text{ScoreSet}_i = \{\text{score}_{ij}\mid 1\le j\le K\}\) and \(\text{TopScores}_i = \text{Top}_{10}(\text{ScoreSet}_i)\). Then
\begin{equation}
\theta_i = \max\big(\tau,\ \text{median}(\text{TopScores}_i) - \lambda\cdot \text{MAD}(\text{TopScores}_i)\big),
\end{equation}
where \(\tau = 0.1\) and \(\lambda = 1.5\) are chosen in our setting. \(\tau\) is the minimum reranker score threshold and \(\lambda\) controls the strictness of filtering by scaling the MAD, a higher \(\lambda\) allows more relaxed filtering, while a lower \(\lambda\) enforces stricter selection. We keep
\begin{equation}
\tilde{D}_i = \{d_{ij} \mid score_{ij} \ge \theta_i\}, \quad |\tilde{D}_i| \le K_{rerank}=3.
\end{equation}

\textbf{Context Construction.}
The final RAG context is created by concatenating filtered snippets:
\begin{equation}
C_i = \text{Concat}\big([\text{[Info~}m\texttt{]}~d_{im}]\big)_{d_{im}\in \tilde{D}_i}
\end{equation}
Here, \(C_i\) denotes the final textual context for query \(q_i\). Each \(d_{im} \in \tilde{D}_i\) is a paragraph-level snippet selected after reranking. The index \(m\) refers to the position of the snippet within the filtered set \(\tilde{D}_i\). We prepend each snippet with a tag \text{[Info \(m\)]}.

\subsection{Dual-Path Generation}
\label{sec:dual-gen}

The primary task of the VLM is to then generate an accurate and reliable answer for each user query \(q_i\). Not all queries have relevant contexts in the knowledge graph, and in some cases, the reranker may assign low relevance scores to all retrieved snippets, resulting in a minimal or empty context \(C_i\). Additionally, the VLM's prior knowledge may still yield correct answers even when relevant external context is unavailable. To balance grounding in retrieved evidence with the model's prior capabilities, we adopt a dual-path generation strategy. Specifically, for each query \(q_i\), we generate two responses: one using retrieval-augmented generation (RAG) and one without context (non-RAG):
\begin{align}
a_i^\prime &= f(q_i, I_i, C_i), \\
a_i &= f(q_i, I_i),
\end{align}
where \(f\) shares the same vision-language model with the summary process. The RAG answer \(a_i^\prime\) is conditioned on the retrieved and filtered context \(C_i\), while the non-RAG answer \(a_i\) is generated without any retrieval and relies on the VLM's prior knowledge only.

\textbf{Self-consistency.} This dual-path setup enables us to later assess the consistency between the two outputs and identify cases where retrieved contexts improve factual grounding. To this end, we adapt the self-consistency check from \cite{wang2023selfconsistency}, with modifications tailored to our dual-path generation setup. After generating answers with and without retrieved context, we evaluate their consistency using an additional verification step. Specifically, we pass the query $q_i$, the retrieved contexts $C_i$ (if any), the RAG answer $a_i^\prime$, and the non-RAG answer $a_i$ into the VLM for comparison.

The motivation follows the self-consistency principle proposed by \cite{wang2023selfconsistency}: LLMs (thus VLMs) (1) are prone to hallucinations without grounding on retrieved contexts, and (2) may produce multiple plausible answers for complex queries. Since relevant context from the knowledge graph is not guaranteed for every query, self-consistency helps ensure that the model produces stable, consistent responses when prompted multiple times. Even when no external context is retrieved, the inherent stochastic nature of VLMs can cause them to generate slightly different answers from the same input. The check thus functions to probe the stability of the model's internal knowledge for answers. In cases where a relevant context is found, agreement between the RAG and non-RAG answers indicates consistency and strengthens confidence in the generated output by showing that it is both grounded and aligned with the LLM's prior knowledge.

To perform this check, the VLM is prompted as an impartial judge tasked with determining whether the two answers are consistent with each other and with the provided context and image. It is instructed to respond with a binary decision: \emph{yes} or \emph{no}. Full prompt details are provided in Appendix~\ref{appendix:self-consistency}.

\begin{table*}[h]
\centering
\caption{Ablation study on Task 1 single-source augmentation under local evaluation.}
\begin{tabular}{lcccc}
\toprule
\textbf{} & \textbf{Accuracy (\%)} & \textbf{Missing Rate (\%)} & \textbf{Hallucination Rate (\%)} & \textbf{Truthfulness Score (\%)} \\
\midrule
LLaMA Vision Only      & 25.00 & 15.38 & 59.62 & -34.62 \\
RAG Agent      & \textbf{27.88} & \textbf{9.62} & 62.50 & -34.62 \\
w/o CoV \& Self-Consistency      & 3.85 & 95.19 & 0.96 & 2.88 \\
w/o CoV      & 4.81 & 95.19 & \textbf{0} & 4.81 \\
Ours    & 14.42 & 82.69 & 2.88 & \textbf{11.54} \\
\bottomrule
\end{tabular}
\label{tab:performance}
\end{table*}

\subsection{Verification and Finalization}
While the dual-path generation and self-consistency check effectively reduce hallucinations, we observe that the system tends to become overly conservative, often defaulting to uncertain or evasive responses (e.g., ``I don’t know'') in borderline cases. This cautious behavior, while safe, may lead to missed opportunities where the answer is in fact correct but filtered out due to lack of full agreement or incomplete grounding. To address this issue and strike a better balance between reliability and informativeness, we introduce a verification-driven finalization step.

For each RAG-generated candidate answer \(a_i^\prime\), we employ a structured ``Chain-of-Verification'' (CoV) protocol \cite{dhuliawala-etal-2024-chain} that systematically probes factual consistency and justification. The process involves two phases: (1) holistic check and (2) decompositional check.

In the holistic check, a verification-oriented VLM evaluates the generated answers against criteria including factual accuracy relative to visual evidence, consistency with provided context, absence of contradictions, direct relevance to the user's query, and specificity. Answers failing in the preliminary check are immediately assigned low confidence scores ($\leq$ 0.5), preventing the propagation of uncertain information. 

For answers passing the initial holistic screening, a decompositional check is performed. The query is broken down into sub-questions \( Q = \{q_1, q_2, \dots, q_n\} \), and each sub-answer \( a_i \) is independently verified. Each verification result provides:

\begin{itemize}
    \item \textbf{Confidence Score} A numeric value between 0.0 and 1.0 reflecting the certainty, where 0.0 indicates complete wrong and 1.0 indicates complete confidence.
    \item \textbf{Reasoning} A brief (1-2 sentence) rationale explaining the reason.
    \item \textbf{Sub-Questions} A concise decomposition of the original question into sub-questions, each accompanied by a finding of ``Supported" or ``Unsupported" based on the verification.
\end{itemize}

To quantitatively set thresholds for answer acceptance, we defined two confidence thresholds: a High Confidence Threshold (1.0), where answers equal to this threshold are deemed highly reliable and directly used, and a Low Confidence Threshold (0.9), where answers falling between this and the high threshold are cautiously accepted if contextually supported and self-consistent. These thresholds were set empirically. The choice of confidence values (0.9 and 1.0) is a direct response to the scoring system of the challenge, which heavily penalizes hallucinations. This approach pragmatically prioritizes the factual accuracy required to perform well under the given evaluation constraints.

This verification-centric approach significantly reduces hallucination rates by implementing multiple validation layers before finalizing any response. Crucially, it not only ensures that our system prioritizes truthfulness over completeness, but also mitigates the overly conservative bias introduced by earlier stages. By assigning calibrated confidence scores, the system is empowered to retain more factually correct answers that might otherwise be discarded, striking a better balance between caution and informativeness. After verification, the system performs a rule-based finalization step to determine whether and which answer to return. This decision process considers multiple signals, including retrieval quality, consistency between RAG and non-RAG answers, and the verifier's confidence score to make a calibrated trade-off between informativeness and reliability. The full decision logic is summarized in Algorithm~\ref{algo:final} below.

\begin{algorithm}
\caption{Final Answer Verification}
\label{algo:final}
\begin{algorithmic}
\REQUIRE \(\, q_i \): query, \(\, C_i \): retrieved context, \(\, a_i^\prime \): RAG answer, \(\, a_i \): non-RAG answer, \(\, S_{\text{CoV}} \): verifier confidence score, \(\, S_{\text{ret}} \): retrieval score, \(\tau_{\text{ret}}\): minimum retrieval quality, \(\tau_{\text{low}}, \tau_{\text{high}}\): verifier confidence thresholds, \texttt{isRealTime}: routing flag
\ENSURE Final answer \( A_i \)

\IF{\texttt{isRealTime} \AND \( S_{\text{ret}} < \tau_{\text{ret}} \)}
    \RETURN ``I don't know''
\ELSIF{\( C_i \neq \emptyset \) \AND IsConsistent\((a_i^\prime, a_i)\) \AND \( S_{\text{CoV}} \geq \tau_{\text{low}} \)}
    \RETURN \( a_i^\prime \)
\ELSIF{\( C_i = \emptyset \) \AND IsConsistent\((a_i^\prime, a_i)\) \AND \( S_{\text{CoV}} \geq \tau_{\text{high}} \)}
    \RETURN \( a_i^\prime \)
\ELSIF{\( C_i \neq \emptyset \) \AND \textbf{not} IsConsistent\((a_i^\prime, a_i)\)}
    \RETURN ``I don't know''
\ELSE
    \RETURN ``I don't know''
\ENDIF
\end{algorithmic}
\end{algorithm}
\section{Experiments}
In this section, we present our experimental setup and evaluation results. To assess the effectiveness of our multimodal RAG pipeline under settings aligned with the CRAG-MM 2025 challenge. All local experiments were conducted on the same hardware as the challenge environment: a single NVIDIA L40S GPU with 48GB of memory. For the generated answer, we employed the GPT-4o-mini~\cite{hurst2024gpt} model as the evaluation LLM.

\subsection{Metrics}
We evaluate our system using the official metric suite defined by the CRAG-MM challenge. Specifically, we report the total number of queries, correct answers, missed answers, and hallucinations. From these, we derive key indicators such as accuracy, hallucination rate, and missing rate. Additionally, we compute the truthfulness score to assess factual grounding.

\subsection{Single-source Augmentation}
To evaluate the effectiveness of our method, we conduct an ablation study on Task 1 under local evaluation, as shown in Table~\ref{tab:performance}. The goal is to assess the incremental impact of the key component in our verification-centric pipeline.

The \textbf{LLaMA Vision Only} baseline refers to the vision-language model (VLM) directly answering questions from image and text input without any external retrieval or augmentation. This setting yields a relatively high accuracy of 25.00\%, but suffers from severe hallucination issues (59.62\%) and a negative truthfulness score of -34.62\%, indicating that the answers often conflict with the ground truth. This is expected, as the VLM attempts to answer all questions regardless of its knowledge limitations, resulting in a low missing rate but poor factual reliability. The \textbf{RAG Agent} baseline incorporates external retrieval via the official search API, enhancing access to relevant knowledge. This improves the accuracy to 27.88\% and slightly reduces the missing rate to 9.62\%. However, hallucination rate increases further to 62.50\%, and the truthfulness score remains at -34.62\%, showing that naive RAG without verification can introduce more misleading content. To further assess the role of verification, we remove key components from our pipeline to observe their individual effects. The \textbf{w/o CoV \& Self-Consistency} variant disables both the self-consistency check and the Chain-of-Verification (CoV). As a result, the model becomes overly conservative, with a missing rate of 95.19\% and very low accuracy (3.85\%). While hallucination rate drops sharply to 0.96\%, the overall coverage and utility of the model are severely degraded. A similar trend is observed in the \textbf{w/o CoV} variant, which enables self-consistency but disables CoV. Although slightly more accurate (4.81\%), it still fails to provide meaningful coverage. Our final agent, denoted as \textbf{Ours}, integrates all proposed components. This configuration strikes the best balance between caution and informativeness. It achieves a truthfulness score of \textbf{11.54\%}, the highest among all systems, with an acceptable hallucination rate of 2.88\% and an improved accuracy of 14.42\%. The results demonstrate that each verification stage contributes to reducing hallucinations while selectively increasing answer confidence, enabling the agent to provide more trustworthy responses.

Our method is developed specifically for the Single-source Augmentation (Task 1) setting. The same pipeline and implementation are directly applied to Task 2 and Task 3 during the CRAG-MM challenge phase. Thus, we focus our discussion on Task 1.
\section{Limitation}
Despite the effectiveness of our proposed system, we encountered several limitations during development and evaluation. First, we initially planned to perform LoRA-based fine-tuning on a vision-language model (VLM) to create a dedicated verifier model tailored to the CRAG-MM task. However, due to limitations of the vLLM framework at the time of the challenge, specifically the version we used did not support loading LoRA adapters after model initialization for LLaMA vision models, we were unable to incorporate LoRA fine-tuned models into our inference pipeline. In addition, the constrained hardware environment provided by the challenge made it infeasible to deploy two VLMs concurrently, which prevented us from experimenting with strategies involving dedicated verification models. Second, we explored fine-tuning on existing VQA datasets to improve factual grounding. However, this approach led to degraded performance, likely due to a distributional gap between standard VQA tasks and the CRAG-MM challenge. The challenge involves complex, multi-source, and temporally grounded queries, often characterized by ambiguous user intent. These scenarios are not well captured by existing VQA benchmarks. As a result, VQA fine-tuning failed to generalize effectively and was ultimately excluded from our final system.
\section{Conclusion}
In this work, we present our solution, a multi-stage, verification-centric RAG framework tailored to the KDD Cup 2025 CRAG-MM challenge. Our system introduces several innovations, including lightweight query routing, dynamic retrieval filtering, dual-path generation, and a structured Chain-of-Verification process. These components work in concert to improve factual consistency, reduce hallucinations, and enhance the reliability of multimodal question answering. Despite encountering practical limitations related to fine-tuning and hardware constraints, our solution demonstrates strong performance in the competition. We believe our pipeline offers generalizable insights for deploying robust, egocentric RAG systems in real-world multimodal settings such as AR/XR environments and smart assistants.

\begin{acks}
This research includes computations using the Wolfpack computational cluster, supported by the School of Computer Science and Engineering at UNSW Sydney. We also acknowledge support from the ARC Centre of Excellence for Automated Decision-Making and Society (CE200100005).
\end{acks}

\bibliographystyle{ACM-Reference-Format}
\bibliography{sample-base}


\appendix

\tcbset{
  mypromptstyle/.style={
    breakable,
    enhanced,
    colback=white,
    colframe=black,
    coltitle=white,
    fonttitle=\bfseries,
    title={#1},
    colbacktitle=gray!60!black,
    boxrule=0.8pt,
    sharp corners,
    before skip=10pt,
    after skip=10pt,
  }
}

\section{Appendix}

\subsection{Prompt Used for Query Routing}
\label{appendix:routing}

\begin{tcolorbox}[mypromptstyle={Query Routing}]
System: You are a query classification expert. Respond concisely with the requested format.
\\ \\
User: Analyze the following user question and classify it. Provide only the classifications. \\ \\ 
Question: "\{query\}" \\ \\
Classifications: \\ \\
1. Needs External Info: [yes/no] (Does answering require knowledge beyond the image?) \\ \\
2. Is Real-Time: [yes/no] (Does it ask about 'today', 'latest', or current events?)
\end{tcolorbox}

\subsection{Prompt Used for Summarizing Image}
\label{appendix:summarizing}

\begin{tcolorbox}[mypromptstyle={Summarizing Image}]
System: You are an expert at describing images with a focus on details relevant to a specific question.
\\ \\
User: \{image\}
A user is asking the following question about the image: '\{query\}'. \\ \\
Please provide a one-sentence summary of the image, focusing on key elements, objects, and any text that might be relevant to answering the question.
\end{tcolorbox}

\subsection{Prompt Used for Self Consistency Check}
\label{appendix:self-consistency}

\begin{tcolorbox}[mypromptstyle={Self Consistency Check}]
System: You are a verification model. Respond with 'yes' or 'no'.
\\ \\
User: \{image\}
You are an impartial judge. You will determine if two answers to a question are consistent with each other and the provided context and image. \\ \\
\#\# Context: \\ \\ \{contexts if contexts else 'No text context.'\} \\ \\
\#\# Question: \\ \\ \{queries\} \\ \\
\#\# Proposed Answer 1: \\ \\ \{answer\} \\ \\
\#\# Proposed Answer 2: \\ \\ \{ragged\_answer\} \\ \\
\#\# Your Task: \\ \\
Are the answers consistent with each other? Respond with only 'yes' or 'no'.
\end{tcolorbox}

\subsection{Prompt Used for Chain-of-Verification}
\label{appendix:cov}
We place the confidence score before the reasoning chain and sub-questions to ensure consistent output and prevent latency issues. This ordering guarantees that every answer is scored, even when generation is truncated due to time or length limits.

\begin{tcolorbox}[mypromptstyle={Chain-of-Verification}]
System: You are a verification expert. Follow the output format strictly.
\\ \\
User: \{image\}
You are an expert fact-checker. Carefully verify if the given answer is correct and well-supported by the image and context. and then decompositionally if it passes the first check. Be very concise. \\ \\

**Crucial Rule:** If the answer requires external knowledge (e.g., names, dates, numbers) that is NOT in the context AND NOT visible in the image, assign a confidence score of 0.0. \\ \\

**Phase 1: Holistic Check.** Quickly assess the answer against these general criteria: \\ \\
1. Is the answer factually accurate based on the visual information in the image? \\ \\
2. If context is provided, is the answer consistent with it? \\ \\
3. Are there any contradictions or uncertain claims? \\ \\
4. Does the answer directly address the user's question? \\ \\
5. Is the answer specific enough to be useful? \\ \\

**If the answer fails Phase 1 check, immediately assign a low confidence score (<= 0.5).** \\ \\

**Phase 2: Decompositional Check (Only if Phase 1 is passed).** \\ \\
1. **Decompose:** Break the question into core sub-questions. \\ \\
2. **Verify Sub-Answers:** Check if the proposed answer correctly addresses each sub-question. \\ \\

Rate your confidence on a scale of 0.0 to 1.0, where: \\ \\
- 1.0 = Completely confident, answer is definitely correct \\ \\
- 0.9 = Very confident, minor uncertainty \\ \\
- 0.7 = Moderately confident, some uncertainty \\ \\
- 0.5 = Low confidence, significant uncertainty \\ \\
- 0.25 = Very low confidence, answer likely incorrect \\ \\
- 0.0 = No confidence, answer is definitely wrong \\ \\

\#\# Context: \\ \\ {contexts if contexts else 'No text context provided.'} \\ \\
\#\# Question:\\ \\ {queries} \\ \\
\#\# Proposed Answer: \\ \\ \{answers\} \\ \\
\#\# Your Task (Follow format strictly): \\ \\
First, provide a confidence score. Then, on a new line, provide a **brief (1-2 sentences)** step-by-step reasoning for your verification. After that, on a new line, provide brief sub-questions and finding. \\ \\
Respond in the following format ONLY: \\ \\
CONFIDENCE: [A single float value between 0.0 and 1.0] \\ \\
REASONING: [Your brief reasoning here] \\ \\
SUB-QUESTIONS: Q1: [Sub-question 1], Finding: [e.g., Supported], Q2: [Sub-question 2], Finding: [e.g., Unsupported]...
\end{tcolorbox}

\end{document}